%% file: main.tex
\documentclass{article}
\usepackage{iclr2026_conference,times}

\input{math_commands.tex}

\usepackage{hyperref}
\usepackage{url}

\usepackage{amsmath}
\usepackage{pifont}
\usepackage{booktabs}
\usepackage{multirow}
\usepackage{makecell}
\usepackage{arydshln}
\usepackage{pgf}
\usepackage{subcaption}

\title{NeRV-Diffusion: Diffuse Implicit Neural \\ Representations for Video Synthesis}

\author{Yixuan Ren, Hanyu Wang, Hao Chen, Bo He \& Abhinav Shrivastava \\
University of Maryland, College Park \\
Project Page: \url{https://nerv-diffusion.github.io/}
}

\iclrfinalcopy
\begin{document}

\maketitle

\input{secs/0_abstract}
\input{secs/1_intro}
\input{secs/2_rw}
\input{secs/3_method}
\input{secs/4_exp}
\input{secs/5_conclusion}

\bibliography{main}
\bibliographystyle{main}

\appendix
\input{secs/X_suppl}

\end{document}

%% file: math_commands.tex
\usepackage{amsmath,amsfonts,bm}

\def\eqref#1{equation~\ref{#1}}

\def\1{\bm{1}}

\DeclareMathAlphabet{\mathsfit}{\encodingdefault}{\sfdefault}{m}{sl}
\SetMathAlphabet{\mathsfit}{bold}{\encodingdefault}{\sfdefault}{bx}{n}



%% file: secs/0_abstract.tex
\begin{abstract}
We present NeRV-Diffusion, an implicit latent video diffusion model that synthesizes videos via generating neural network weights.
The generated weights can be rearranged as the parameters of a convolutional neural network, which forms an implicit neural representation (INR), and decodes into videos with frame indices as the input.
Our framework consists of two stages:
1) A hypernetwork-based tokenizer that encodes raw videos from pixel space to neural parameter space, where the bottleneck latent serves as INR weights to decode.
2) An implicit diffusion transformer that denoises on the latent INR weights.
In contrast to traditional video tokenizers that encode videos into frame-wise feature maps, NeRV-Diffusion compresses and generates a video holistically as a unified neural network.
This enables efficient and high-quality video synthesis via obviating temporal cross-frame attentions in the denoiser and decoding video latent with dedicated decoders.
To achieve Gaussian-distributed INR weights with high expressiveness, we reuse the bottleneck latent across all NeRV layers, as well as reform its weight assignment, upsampling connection and input coordinates.
We also introduce SNR-adaptive loss weighting and scheduled sampling for effective training of the implicit diffusion model.
NeRV-Diffusion reaches superior video generation quality over previous INR-based models and comparable performance to most recent state-of-the-art non-implicit models on real-world video benchmarks including UCF-101 and Kinetics-600.
It also brings a smooth INR weight space that facilitates seamless interpolations between frames or videos.
\end{abstract}

%% file: secs/1_intro.tex
\section{Introduction}
\label{sec:intro}

Video latent diffusion models (LDMs) have achieved impressive generative capability.
However, their tokenizers usually inherit from those of image diffusion models and encode videos as individual frame-wise feature maps, ignoring the natural coherence across frames and resulting in redundant representations.
Cross-frame attentions \citep{wang2023modelscope,guo2023animatediff} are thus introduced to constrain temporal consistency in both generation and decoding processes, largely increasing the model size and leading to massive computation footprint.
Moreover, a typical tokenizer for diffusion \citep{rombach2022high} is usually built in the form of a large-scale variational autoencoder (VAE), which compresses the visual data into latent code with generalizable decoding quality on diverse data.
During inference, the denoised latent must be processed by the decoder to be rendered into pixels, demanding high computation for visualization efficiency.

Implicit neural representations (INRs) are neural networks that fit on single data points.
An INR takes unified coordinates as the input and outputs pixels as stored in its model weights.
It has shown significant advantages on compression \citep{sitzmann2020implicit, dupont2021coin}, fast decoding \citep{chen2021nerv}, and easy transformation \citep{mildenhall2021nerf, kerbl20233d} by representing data as an integral format of function.
The continuity and differentiability of INRs facilitate advanced single-data generative tasks, such as super-resolution, restoration, style transfer and editing, via smooth interpolations within the data space.
Its compact representation also contributes to reducing memory overhead, making them highly suitable in resource-constrained environments.

\begin{figure}[t]
    \centering
    \begin{subfigure}[b]{0.59\linewidth}
    \centering
    \includegraphics[width=\linewidth]{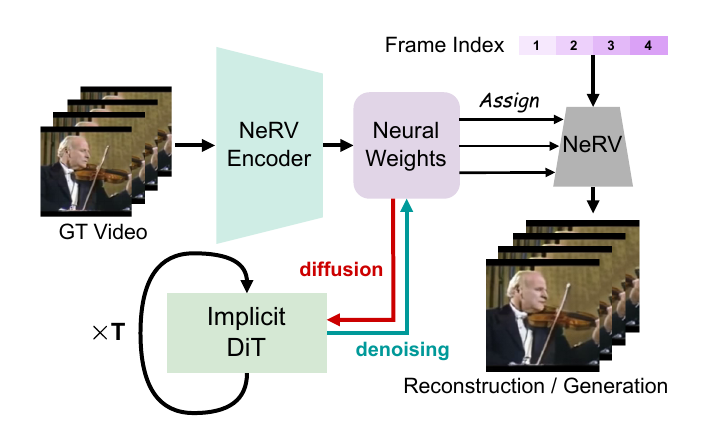}
	\end{subfigure} \hfill
	\begin{subfigure}[b]{0.4\linewidth}
    \centering
    \includegraphics[width=\linewidth]{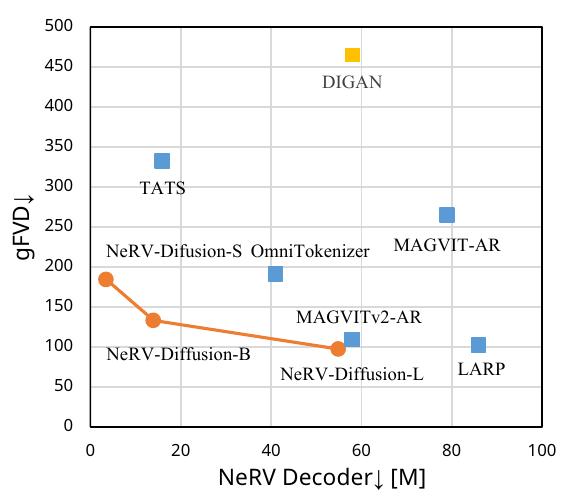}
	\end{subfigure}
    \caption{
    \textbf{Left:} Overview of our NeRV-Diffusion framework.
    In the tokenization stage, NeRV encoder projects RGB videos to neural network weights, forming up NeRVs and decoding for reconstruction.
    In the generation stage, an implicit diffusion transformer is trained to denoise on NeRV weights.
    During inference, the implicit DiT generates NeRV weights from random noise, which decode into RGB videos.
    \textbf{Right:} NeRV-Diffusion (\textcolor[HTML]{ed7d31}{orange}) outperforms previous INR-based (\textcolor[HTML]{ffc000}{yellow})  as well as most recent non-implicit (\textcolor[HTML]{5b9bd5}{blue}) video generation models at all scales with more compact model sizes.
    The generative performance is evaluated in gFVD on UCF.
    }
    \label{fig:teaser}
\end{figure}

To harness the strengths of both latent generative models and implicit neural representations, we establish an implicit latent diffusion model, NeRV-Diffusion, for video synthesis by generating INR weights, where a video is represented as a holistic set of INR weights.
It consists of two stages:
In the tokenization stage, a hypernetwork-based encoder compresses RGB videos into parametric latent tokens.
The tokens instantiate an INR to decode for reconstruction with unified frame indices input.
In the generation stage, a diffusion transformer denoises in the encoded implicit latent space, mapping random noise to INR weight tokens.
Figure \ref{fig:teaser} (left) overviews the framework.

\begin{figure}[t]
    \centering
    \includegraphics[width=0.9\linewidth]{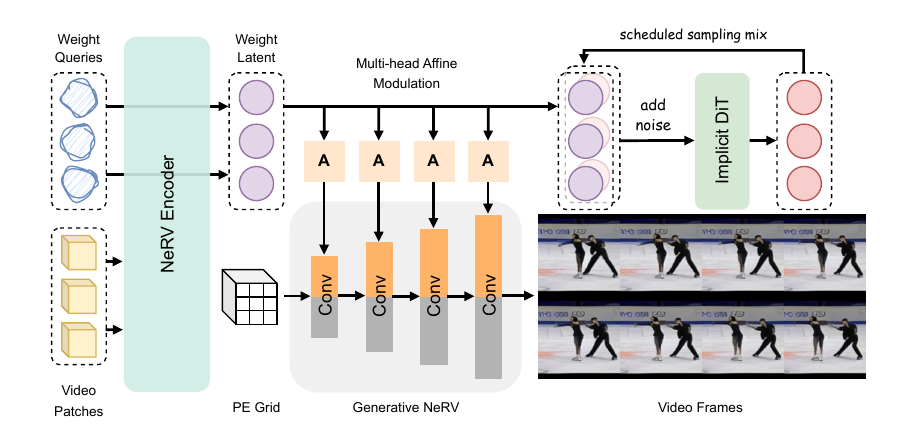}
    \caption{Detailed architectures of NeRV-Diffusion.
    \textbf{Left}: Patchified videos and initialized weight queries are concatenated and input into NeRV encoder, outputting latent weight tokens;
    \textbf{Middle top:} Weight tokens are reused and converted by multi-head affines to instantiate each generative NeRV layer;
    \textbf{Middle bottom:} Generative NeRV decodes spatiotemporal positional embeddings into RGB videos, using the instance-specific modulation weights (\textcolor[HTML]{FFB302}{gold}) and global shared weights (\textcolor[HTML]{555555}{gray}).
    Block details and side connections are omitted;
    \textbf{Right:} Weight tokens are added noise and an implicit diffusion transformer is trained to denoise in this implicit weight space.
    }
    \label{fig:archs}
\end{figure}

However, it is not trivial to acquire Gaussian-distributed neural network weights for smooth diffusion that are meanwhile able to represent diverse realistic data with high fidelity.
We adopt a convolutional video INR, NeRV \citep{chen2021nerv}, and build a transformer INR encoder based on FastNeRV \citep{chen2025fast}.
They are originally designed toward video compression performance only and their produced INR weights are not generatable.
To ensure the bottleneck latent tokens fitful for both faithful reconstruction and smooth diffusive generation, we have made several critical architectural modifications.
The detailed architectures are illustrated in Figure \ref{fig:archs}.

Specifically, we reuse the encoded weight tokens with multiple linear affine layers such that each NeRV layer is modulated by all tokens independently.
We also redesign the weight modulation approach, proposing to directly set the latent tokens to be the convolution kernels, instead of repeating and multiplying them with shared base weights.
These upgrades fundamentally enlarge the expressiveness and smoothness of the implicit space while maintaining its compactness.
We leverage vanilla diffusion transformer \citep{peebles2023scalable} (DiT) to denoise on weight tokens that imply no spatial or temporal structures.
We also handle the error accumulation with SNR-adaptive loss weighting and scheduled sampling for optimal denoising in the implicit latent space.

NeRV-Diffusion leverages video INRs as instance-specific decoders, offering faithful reconstruction, compact model and fast decoding compared to the large, shared decoders in traditional LDMs.
It encodes and generates video frames holistically as integral INR weights, implies the keyframe-residue representation by reusing the same set of parameters to decode all frames, and thus maintains temporal associations without cross-frame attention.
Furthermore, NeRV-Diffusion generates neural weights with only a single linear layer after the Gaussian bottleneck, and employs direct channel-wise parameterization to construct the INRs.
This leads to multi-variant normal distribution of our generative NeRV weights and enables smooth interpolation between frames and videos.

In summary, our contributions are as follows:
\begin{itemize}
    \item We propose a novel implicit video autoencoder that compresses videos into neural weight tokens of normal distribution, constituting generation-specialized video INRs.
    \item We propose an implicit diffusion model that denoises in neural weight space, achieving dynamic and diverse video synthesis via generating INR parameters.
    \item NeRV-Diffusion surpasses prior implicit and most recent non-implicit generative models on multiple real-world video benchmarks, and conveys smooth time and weight interpolations.
\end{itemize}

%% file: secs/2_rw.tex
\section{Related Work}

\subsection{Implicit Neural Representations}

Implicit neural representations (INRs) are neural networks that fit on single data points.
An INR takes in coordinates and outputs corresponding pixel values of the stored data.
It has presented capacity and flexibility in various modalities, including images \citep{sitzmann2020implicit, dupont2021coin}, 3D shapes \citep{park2019deepsdf, mildenhall2021nerf} and videos \citep{chen2021nerv,chen2022videoinr}.
They are primarily developed for image compression \citep{strumpler2022implicit,dupont2022coin++} and editing \citep{fan2022unified,yang2023implicit}, video compression \citep{li2022nerv,kwan2024hinerv,zhao2023dnerv,zhang2021implicit,lee2023ffnerv} and editing \citep{ouyang2024codef}, and novel view rendering \citep{kerbl20233d,barron2023zip,cao2023hexplane} and 3D scene editing \citep{yuan2022nerf,liu2024genn2n}.
Although some editing applications have been explored, they create the INRs after manipulating the data in pixel space.

A standard INR is trained via memorizing the pixel data, which is time-consuming in a backpropagation manner.
\cite{chen2022transformers,kim2023generalizable,chen2025fast} suggest using transformer-based hypernetworks to create INR weights given RGB data in a feed-forward fashion at scale.
However, these methods are optimized solely toward reconstruction performance and incorporate no distribution regularization on the produced INR weights, leaving the implicit generative task that synthesizes novel data points from random noise under-addressed.

\subsection{Implicit Neural Representation Generation}

INR generation is a challenging task.
Traditional generative models learn mapping random noise to pixels or latent features, while implicit generative models aim to associate neural parameters with Gaussian distribution.
Several efforts have been made toward implicit generation.
\cite{skorokhodov2021adversarial} builds a GAN for image INRs \citep{sitzmann2020implicit}, and \cite{yu2022generating} extends it to videos by involving the temporal axis.
\cite{erkocc2023hyperdiffusion,chen2023single,muller2023diffrf,shue20233d} study generating 3D NeRF parameters via diffusion models.
\citep{chen2024image} applies latent diffusion models on image INRs \citep{chen2021learning} for image synthesis, while their INR weights are derived by a complex decoder from the denoising latent space.
Recently, \cite{wang2024inrflow,wang2025pixnerd} propose to leverage the hypernetwork-INR architecture to conduct flow matching on image or 3D pixel data.
\cite{lee2025minr} also developed a masked image autoencoder for inpainting with a similar structure.
Despite these efforts, no video diffusion model that generates INR weights has yet been explored, casting this a challenging task as videos embed more dynamic information and diffusion models have a more strict demand on its denoising space.

\subsection{Latent Video Diffusion Models}

Latent video diffusion models \citep{wang2023modelscope,blattmann2023align,guo2023animatediff} have achieved significant success in video generative modeling.
However, traditional video tokenizers often encode video frames as individual feature maps, calling cross-frame attentions in the denoising network to constrain temporal consistency.
\cite{kim2023diffusion,wu2025improved} start to explore video autoencoders with motion awareness and temporal compression, splitting the complexity between the tokenization and generation stages.
Recent 1D tokenization \citep{yu2024image,wang2024larp,zha2025language} encodes visual data into holistic tokens that project no spatial or temporal alignment with pixels, while they remain focused on images or auto-regressive generation only.
In this work, we look to synthesize videos by generating INR weights via diffusion, obviating frame-wise representations by using the whole INR model to decode all frames given time indices.
Moreover, symmetric autoencoders rely on a large-scale decoder to render synthesized latent to diverse RGB data with high fidelity, consuming non-negligible computational resources and time for end users to visualize.
We explore the space of asymmetric hypernetwork-INR autoencoders, where the INR acts as an efficient instance-specific decoder as it only needs to represent a single data point.

%% file: secs/3_method.tex
\section{NeRV Diffusion}
\label{sec:method}

NeRV-Diffusion is a two-stage generative framework.
In the tokenization stage, an implicit autoencoder (\S\ref{sec:hypernerv}) is trained to compress a video from pixels to latent neural weight tokens, and the tokens function as the parameters of an INR (\S\ref{sec:gnerv}) and self-decode to reconstruct the video.
In the generation stage, an implicit diffusion transformer (\S\ref{sec:nervdit}) is trained to generate the weight tokens from random noise.
Figures \ref{fig:teaser} (left) and \ref{fig:archs} illustrate our full pipeline of both stages.

\subsection{NeRV Autoencoder}
\label{sec:hypernerv}

In the first stage, we aim to tokenize a video into a latent space that represents the video through the parameters of an INR.
This is achieved by training an implicit autoencoder, where the encoder $\mathcal{E}$ is a hypernetwork that produces INR parameters $\theta=\mathcal{E}(x)$ given pixel input $x$.
The decoder is implemented as an INR $\mathcal{D_{\theta}}(\cdot)$, which decodes to pixel values given corresponding coordinates.
We build the backbone of our INR encoder $\mathcal{E}$ upon ViT-based FastNeRV \citep{chen2025fast}, where we make several critical modifications to align the learned latent space with generative tasks.

The RGB video is first segmented into patches and converted to transformer input embeddings.
Since the output weight tokens have no spatiotemporal correspondence to the input patches, instead of mapping them directly we introduce dedicated query tokens and concatenate them with the data patches following \citep{peebles2022learning}.
Only the output tokens corresponding to the queries are retained.
They are batch normalized along the token embedding dimension.

\paragraph{KL Bottleneck.}
Two additional fully connected (FC) layers are appended after the NeRV encoder's output to create an information bottleneck of compact latent dimension.
KL divergence loss is applied to align their distribution toward standard Gaussian distribution $\mathcal{N}(0, 1)$.

\begin{figure}[t]
    \centering
    \includegraphics[width=0.92\linewidth]{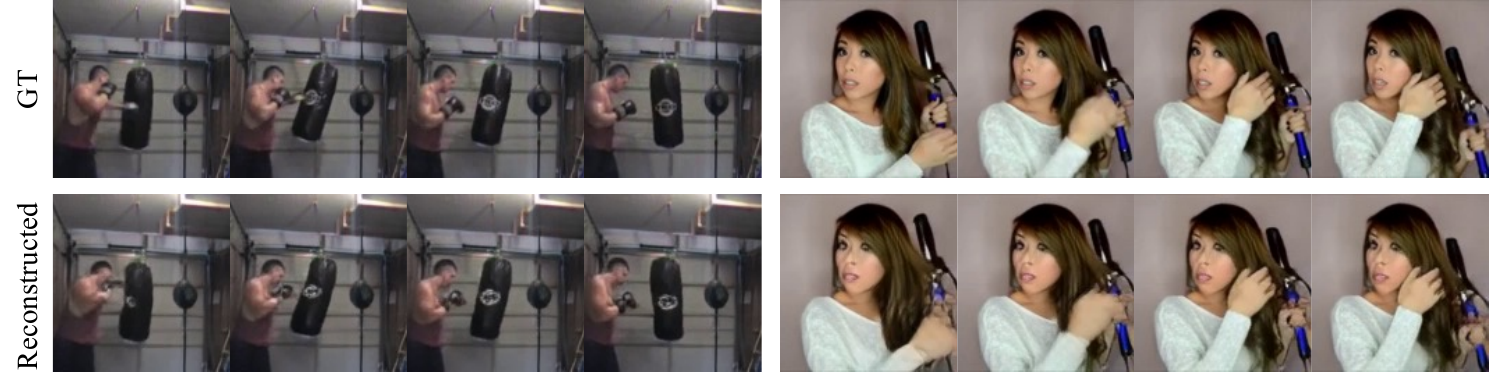}
    \caption{Video reconstruction of our NeRV autoencoder on UCF (left) and K600 (right).
    }
    \label{fig:rec}
\end{figure}

\paragraph{Multi-head Affine Mapping.}
FastNeRV use its encoded latent to modulate the parameters of a subset of the INR layers, which limits the capacity of the latent tokens especially when KL constraint is applied for generation tasks.
Inspired by the multiple affine layers in \cite{karras2019style}, we expand the post-bottleneck FC layer into multi-head affine mappings, and the single set of weight tokens are reused to modulate all NeRV layers independently.
Specifically, for each NeRV layer, a dedicated affine head maps all the weight tokens into modulation parameters.
This strategy significantly expands the expressiveness of the weight tokens, as a compact latent space will reduce the complexity of the diffusion process in the generation stage.

\paragraph{Channel-wise INR Parameterization.}
FastNeRV repeats the weight tokens and multiply them to the instance-agnostic INR base weights via dot product as the modulation.
\cite{skorokhodov2021adversarial,yu2022generating} perform low-rank vector cross product to amplify the modulation matrix dimension from condensed weight latent.
Inspired by \cite{lin2021anycost} that prunes a pretrained GAN generator by subsetting its kernels, we propose to directly set affined instance-specific weight tokens to be the convolutional kernels at a certain group of INR channels.
Other parameters $\theta_{s}$ are shared among all training data and are learnable during training.
All kernel values are normalized along all dimensions except the output channels, following the demodulation in \cite{karras2019style}.
In this way, the generated weight tokens are directly involved in decoding with maximal degrees of freedom.
This also enables smooth parameter interpolation between our INR decoders.

\paragraph{Convolutional Discriminator.}
To generate realistic videos we incorporate adversarial training \citep{goodfellow2020generative}.
We choose a convolutional discriminator \citep{karras2019style} over a transformer-based one, as we observe that the latter introduces flickering artifacts across frames.

\paragraph{Training Objectives.}
We train NeRV-VAE with the reconstruction objective.
With an additional perceptual loss \citep{zhang2018unreasonable} $\mathcal{L}_{\mathrm{LPIPS}}$ and the adversarial loss $\mathcal{L}_{\mathrm{GAN}}$,
it is optimized via
\begin{equation}
    \mathcal{L}_{\mathrm{VAE}}(\mathcal{E}, \theta_{s}) =
    \|x - \tilde{x}\|^{2} +
    \mathcal{L}_{\mathrm{LPIPS}}(x, \tilde{x}) +
    \mathcal{L}_{\mathrm{GAN}}(x, \tilde{x}) +
    D_{\mathrm{KL}}(\mathcal{N}(0, 1), \tilde{\theta}).
\label{eq:vae}
\end{equation}

\subsection{Generative NeRV}
\label{sec:gnerv}

The encoded weight tokens are formed into a video INR $\mathcal{D_{\theta}}(\cdot)$ that decodes to reconstruct the video.
NeRV \citep{chen2021nerv} is a convolutional video INR that takes time index $t$ as the input query and yields a whole frame at each forwarding.
We construct our implicit decoder based on it while introducing several upgrades to enhance its capacity for generative purposes.

\paragraph{Spatiotemporal Embedding Input.}
Time-query video INRs upsamples from $\mathbb{R}^{T\times D\times1\times1}$ to $\mathbb{R}^{T\times 3\times H\times W}$, where no spatial dimension is input.
With this structure, we observe distinct movement in the reconstruction, however the spatial content lacks clarity.
To balance between the appearance and motion quality, we expand the input time embedding to 3D spatiotemporal, while time remains the sole query axis.
Specifically, we sample a 3D positional embedding and reshape it to $ \mathbb{R}^{T\times 3D\times h\times w}$.
This spatiotemporal input supplements geometric prior and avoids the leading FC layer in vanilla NeRV that were designed for transforming 1D time embedding input.
We observe that full convolutions fit optimally for generative quality with our multi-head affine modulation.

\paragraph{Scaling up Blocks.}
Benefited from the reused weight modulation with multi-head affine mappings, we are able to largely scale up our generative NeRV without extra weight tokens.
We expand the upsampling layers to blocks, each performing one-level ($2\times$) upsampling with additional convolutions that don't change the shape.
Compared to the assorted upsampling scales in limited layers in vanilla NeRV, this periodic upsampling structure evenly distributes the information from low to high resolutions, and cooperates well with our multi-head affine modulation.
We also double the hidden dimensions of the layers in the last block following \cite{karras2020analyzing} so that more native high-resolution information can be processed with sufficient capacity.

\paragraph{Upsampling Algorithm.}
While vanilla NeRV has tested that pixelshuffle results in the best reconstruction performance with similar amount of parameters, we again compare different upsampling algorithms for our generative NeRV.
We find that transposed convolutions achieve non-negligible better generation quality to pixelshuffle with merely a quarter of parameters and computations.
Therefore we choose transposed convolutions for all the upsampling layers in our generative NeRV.

\paragraph{Side Connections.}
With the increased depth of our generative NeRV by upscaled blocks, we further append side connections to effectively collate all intermediate resolution information with minimal computation overhead.
We investigate the residual and skip connections as in \cite{karras2020analyzing}.
If the side connection needs additional layers, they are also modulated by the same set of our weight tokens thanks to our multi-head affine mappings and no extra trainable parameter is introduced.
Residual connection fuses latent features at different scales before decoding to RGB and is experimented to yield clearer appearance and stabler motion.

\begin{table}[t]
    \centering
    \caption{Model and bottleneck representation size comparison.
    rFVD and gFVD are results on UCF.
    $^{\dagger}$ Note that for implicit GANs we conceptually separate the mapping network as the generator and the generator network as the decoder, as the latter takes in the frame index and decodes as the INR.}
    \label{tab:gen}
    \resizebox{\linewidth}{!}{
        \begin{tabular}{lccccccc}
        \toprule
        \textbf{Method} & \multicolumn{2}{c}{\textbf{\#Params}} & \textbf{\#Tokens} & \multicolumn{2}{c}{\textbf{rFVD}$\downarrow$} & \multicolumn{2}{c}{\textbf{gFVD}$\downarrow$} \\
         \cmidrule{2-3} \cmidrule{5-6} \cmidrule{7-8}
         & Detokenizer & Generator & & UCF & K600 & UCF & K600 \\
        \midrule
        \textit{Non-Implicit Models} \\
        \hdashline
        \addlinespace
        CogVideo \citep{hong2022cogvideo} & - & $9.4$B & - & - & - & $626$ & $109$ \\
        TATS \citep{ge2022long} & $16$M & $362$M & $1024$ & $162$ & - & $332$ & - \\
        MAGVIT-AR \citep{yu2023magvit} & $79$M & $306$M & $1024$ & $25$ & - & $265$ & - \\
        Latte \citep{ma2024latte} & $49$M & $674$M & $512$ & $21$ & - & $202$ & - \\
        OmniTokenizer \citep{wang2024omnitokenizer} & $41$M & $650$M & $1280$ & $42$ & - & $191$ & $33$ \\
        VideoFusion \citep{luo2023videofusion} & - & $2$B & - & - & - & $173$ & - \\
        MAGVITv2-AR \citep{yu2023language} & $58$M & $840$M & $1280$ & $\mathbf{8.6}$ & - & $109$ & - \\
        LARP \citep{wang2024larp} & $86$M & $343$M & $1024$ & $\underline{20}$ & - & $\underline{102}$ & $\mathbf{6.2}$\\
        \midrule
        \textit{Implicit Models} \\
        \hdashline 
        \addlinespace
        DIGAN \citep{yu2022generating} & $58$M$^{\dagger}$ & $5.5$M$^{\dagger}$ & - & - & - & $465$ & - \\
        NeRV-Diffusion-S (Ours) & $\mathbf{3.5}$M & $467$M & $128$ & $85$ & $40$ & $184$ & $46$ \\
        NeRV-Diffusion-B (Ours) & $\underline{14}$M & $467$M & $128$ & $59$ & $27$ & $133$ & $30$ \\
        NeRV-Diffusion-L (Ours) & $55$M & $467$M & $128$ & $41$ & $19$ & $\mathbf{97}$ & $\underline{22}$ \\
        \bottomrule
        \end{tabular}
    }
\end{table}

\subsection{Implicit Diffusion}
\label{sec:nervdit}

With visual data tokenized from pixel space to NeRV weight space by the implicit autoencoder described above, we perform diffusion process on these weight tokens by $\theta_{t}=\alpha_{t}\theta_{0}+\sigma_{t}\epsilon$ and train a denoising network $\phi$ toward
\begin{equation}
    \mathcal{L}_{\mathrm{IDM}} = E_{\theta, \epsilon\sim \mathcal{N}(0,1), t} [ \| \epsilon_{0} - \epsilon_{}(\epsilon_{t}, t) \|^{2} ]
\end{equation}

It is not trivial to model denoising process on neural weights.
Previous diffusion models are designed for pixel data or their latent feature maps.
Since NeRV weight tokens have no spatiotemporal structure, transformers are more suitable than U-Nets to process them, and temporal attention is unnecessary in our denoising network like those in traditional video diffusion models \citep{ma2024latte}.
G.pt \citep{peebles2022learning} uses transformers to evolve neural network weights in a meta-learning fashion but not on noisy data.
DiT \citep{peebles2023scalable} tailors transformers for image diffusion and \cite{zha2025language} also use it to process 1D image tokens in diffusion.
We explored these backbone options and DiT reaches the optimal performance with a straightforward architecture.
Besides, we curate the training scheme as below to fill the gap when adapting DiT to the implicit space.

\paragraph{Min-SNR-$\gamma$ Loss Weighting.}
We observe that our implicit diffusion model converge slower on early denoising timesteps than late ones, i.e. it is tougher to learn to parse more noisy input.
To address this issue and speed up its convergence, we adopt Min-SNR-$\gamma$ loss weighting \citep{hang2023efficient} and apply the coefficient $w_{t} = \min \{ \mathrm{SNR}(t), \gamma \}$ on the denoising loss, where $\mathrm{SNR}(t) = \frac{\alpha_{t}^2}{\sigma_{t}^{2}}$ reflects the signal-noise ratio at timestep $t$, and constant $\gamma$ controls the minimum of $w_{t}$.
This loss weighting strategy prevent the diffusion training to focus too much on the low noise levels and descends evenly toward the denoising directions at all timesteps.

\paragraph{Scheduled Sampling.}
To further enhance the implicit denoising chain and tackle the exposure bias issue, we introduce scheduled sampling \citep{bengio2015scheduled} into our training scheme.
It is initially proposed for auto-regressive models, and has been applied on diffusion models \citep{ning2023elucidating,ren2024multi} to fill the training-inference gap brought by Teacher Forcing.
During training, after the first forward round at step $t$, we randomly use the model predictions $\tilde{\theta_{t-1}}=\theta_{\phi}(\theta_{t}, t)$ as the new input and execute another forward pass, and calculate the total losses.
It aligns the the training and inference modes, ensures low input disparity and minimizes error accumulation during sampling.

%% file: secs/4_exp.tex
\section{Experiments}

\subsection{Setups}
\label{sec:setup}

\paragraph{Datasets.}
We demonstrate NeRV-Diffusion on two real-world video benchmarks: video generation on UCF-101 \citep{soomro2012ucf101} (UCF) and frame prediction on Kinetics-600 \citep{kay2017kinetics} (K600).
All experiments are conducted on 16 frames of $128^{2}$ resolution.
We use the train split of K600, and all videos from UCF, following prior work \citep{yu2023magvit,wang2024larp}.

\paragraph{Implementations.}
We realize our NeRV encoder with the backbone of Vision Transformer \citep{dosovitskiy2021an} (ViT).
We scale up our generative NeRV decoder to three configurations of progressive sizes: -Small ($3.5$M), -Base ($14$M) and -Large ($55$M).
We ablate our key design options in \S\ref{sec:ablation}.
Detailed model and training configurations are provided in Appendix \ref{sec:suppl_implem}.

\paragraph{Metrics.}
We measure Fr\'{e}chet Video Distance (FVD) \citep{unterthiner2019fvd} to evaluate the reconstruction and generation quality of NeRV-Diffusion.
We calculate FVD on $\pgfmathprintnumber{2048}$ sampled videos following prior work \citep{yu2022generating,yu2023magvit,wang2024larp} for fair comparison.

\begin{figure}[t]
    \centering
    \includegraphics[width=\linewidth]{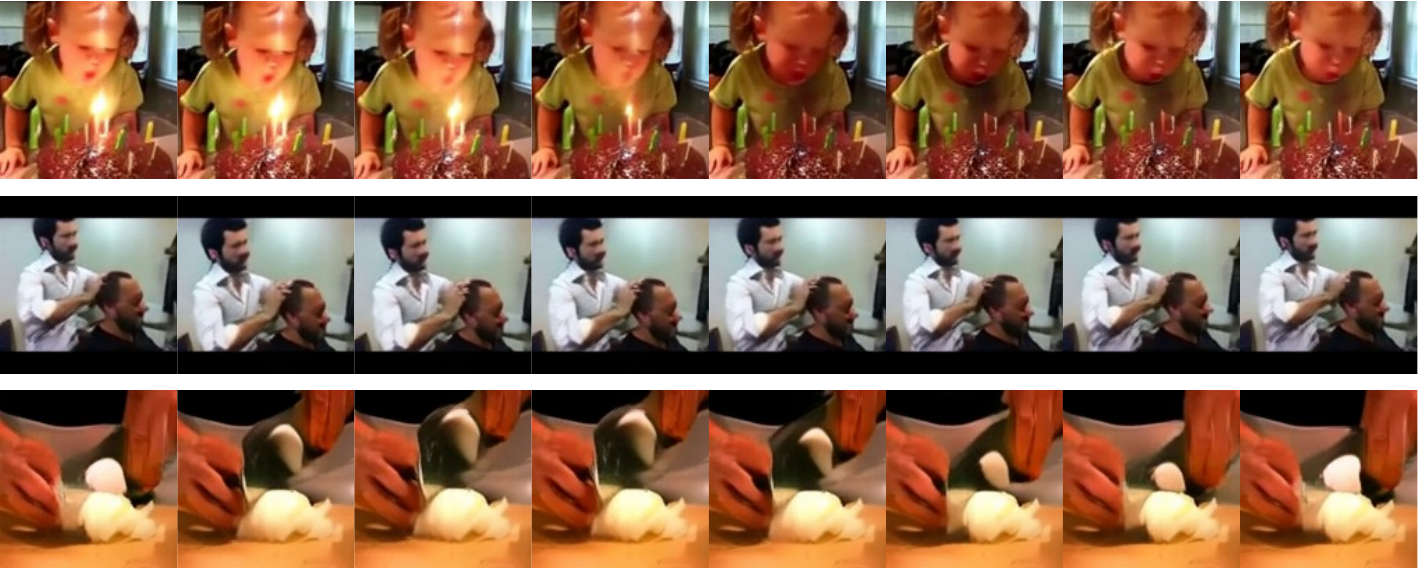}
    \caption{Class-conditioned video generation on UCF.
    }
    \label{fig:ucf_cond}
\end{figure}

\subsection{Video Reconstruction}

Visualized reconstruction output of our implicit tokenizer are displayed in Figure \ref{fig:rec}.
The quantitative results, together with the model and bottleneck latent size comparisons are listed in Table \ref{tab:gen}.
NeRV-Diffusion achieves comparable performance to other non-implicit methods, with much more compact model and latent sizes.
It also worth noting that NeRV-Diffusion features a small reconstruction-generation gap compared to other models, indicating our effective design of implicit video representations for generation purposes, and thus efficient usage of our latent space.

\subsection{Video Generation}

\paragraph{Class-Conditioned Video Generation.}
We conduct class-conditioned video generation on UCF and present our visual results in Figure \ref{fig:ucf_cond}.
We quantitatively compare NeRV-Diffusion with other models in Table \ref{tab:gen}.
NeRV-Diffusion outperforms previous INR-based generative methods as well as most recent non-implicit models of various mechanisms, including GAN, diffusion and auto-regressive architectures.
It is able to synthesize dynamic videos with diversity in both appearances and motions, ranging from detailed objects to complex scenes.
More samples in Appendix \ref{sec:supp_exp}.

\begin{figure}[t]
    \centering
    \includegraphics[width=\linewidth]{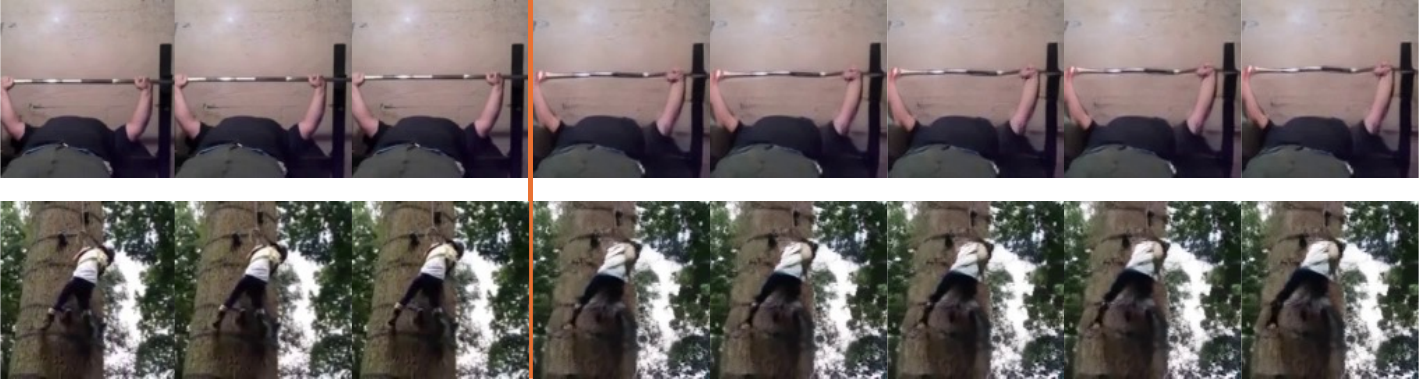}
    \caption{Frame prediction on K600.
    Frames in front of the orange line are input conditions.
    }
    \label{fig:k600}
\end{figure}

\paragraph{Frame Prediction.}
Following the settings in \cite{hong2022cogvideo}, we train our implicit diffusion model given the initial 5 frames to predict the rest 11 frames.
We construct a sequence of the 5 given frames and 11 zero-valued frames and encode them as a video clip to the NeRV weight space.
Its weight tokens are concatenated as the input condition with the noised ground truth.
We also adapt the classifier-free guidance \citep{ho2022classifier} (CFG) to the frame condition.
The quantitative results are listed in Table \ref{tab:gen} and the visualizations are displayed in Figure \ref{fig:k600}.
Our model faithfully propagates the spatial content and movement flows to future frames.

\subsection{Ablation Studies}
\label{sec:ablation}

We conduct ablation studies to assess the key components we propose in \S\ref{sec:method} and validate the optimal design options for generative objectives.
The quantitative results are tested with NeRV-Diffusion-S configuration on UCF and are listed in Tables \ref{tab:ablat_rec} and \ref{tab:ablat_gen}.

Table \ref{tab:ablat_rec_a} indicates that in our NeRV autoencoder, our channel-wise parameterization outperforms due to its maximal transparency to decode directly using the encoded weight tokens.
In Table \ref{tab:ablat_rec_b}, our multi-head affines significantly boost the capacity of NeRV by mapping the whole bottleneck weight tokens to different NeRV layers for reused modulation.
Table \ref{tab:ablat_rec_c} demonstrates that the spatiotemporal input embedding of shape $h=w=8$ expands the input space with peak expressiveness, while smaller sizes lead to truncated space and bigger sizes result in fewer upsampling layers.
We compare different upsampling operations in Table \ref{tab:ablat_rec_d}, and find that transposed convolution surpasses pixelshuffle by much fewer parameters and computations without inflated channels.
We further explore side connection types in Table \ref{tab:ablat_rec_e}, and observe that residual connections fuse raw features at diverse scales without visible artifacts brought by skip connections when summing up multi-resolution RGB output.
Finally we scale up our implicit latent space by increasing the number of tokens, as we meanwhile observe that the token dimension only makes slight impact on the output quality.
$128$ tokens reaches the peak performance and more tokens will lead to an over complex latent space for diffusion although the reconstruction error continues dropping.

For our implicit denoising network, we consider three backbone candidates in Table \ref{tab:ablat_gen_a}.
G.pt was designed for neural weight evolution, but not adapted to diffusion tasks.
Latte was designed for video generation and incorporate with temporal attentions, which are not beneficial to our implicit generation as NeRV weight tokens lack spatiotemporal structure.
Table \ref{tab:ablat_gen_b} showcases that both Min-SNR-$\gamma$ loss weighting and scheduled sampling scheme effectively minimize the gap between implicit diffusion training and inference, by emphasizing the denoising model more on high-noise predictions and imperfect input.

\begin{table}[t]
    \centering
    \caption{Ablation studies of the key design options in our NeRV autoencoder and generative NeRV, tested with NeRV-Diffusion-S and the best implicit denoiser configuration on UCF.}
    \label{tab:ablat_rec}
    \begin{subtable}[t]{0.3\linewidth}
        \centering
        \begin{tabular}{lc}
            \toprule
            Modulation & gFVD$\downarrow$ \\
            \midrule
            Repeat & $741$ \\
            FMM & $636$ \\
            Channel & $\mathbf{570}$ \\
            \bottomrule
        \end{tabular}
        \caption{}
        \label{tab:ablat_rec_a}
    \end{subtable}
    \hfill
    \begin{subtable}[t]{0.3\linewidth}
        \centering
        \begin{tabular}{lc}
            \toprule
            Reuse & gFVD$\downarrow$ \\
            \midrule
            No reuse & $570$ \\
            Direct reuse & $562$ \\
            Multi-head affines & $\mathbf{283}$ \\
            \bottomrule
        \end{tabular}
        \caption{}
        \label{tab:ablat_rec_b}
    \end{subtable}
    \hfill
    \begin{subtable}[t]{0.3\linewidth}
        \centering
        \begin{tabular}{lc}
            \toprule
            Spatial PE & gFVD$\downarrow$ \\
            \midrule
            $h=w=1$ & $283$ \\
            $h=w=4$ & $269$ \\
            $h=w=8$ & $\mathbf{254}$ \\
            $h=w=16$ & $277$ \\
            \bottomrule
        \end{tabular}
        \caption{}
        \label{tab:ablat_rec_c}
    \end{subtable}
    \begin{subtable}[t]{0.3\linewidth}
        \centering
        \begin{tabular}{lc}
            \toprule
            Upsampling & gFVD$\downarrow$ \\
            \midrule
            PixelShuffle & $254$ \\
            Transposed Conv & $\mathbf{248}$ \\
            Bilinear & $287$ \\
            \bottomrule
        \end{tabular}
        \caption{}
        \label{tab:ablat_rec_d}
    \end{subtable}
    \hfill
    \begin{subtable}[t]{0.3\linewidth}
        \centering
        \begin{tabular}{lc}
            \toprule
            Side Connection & gFVD$\downarrow$ \\
            \midrule
            Vanilla & $248$ \\
            Residual & $\mathbf{219}$ \\
            Skips & $235$ \\
            \bottomrule
        \end{tabular}
        \caption{}
        \label{tab:ablat_rec_e}
    \end{subtable}
    \hfill
    \begin{subtable}[t]{0.3\linewidth}
        \centering
        \begin{tabular}{lc}
            \toprule
            Token Shape & gFVD$\downarrow$ \\
            \midrule
            $32\times128$ & $219$ \\
            $64\times128$ & $193$ \\
            $128\times128$ & $\mathbf{184}$ \\
            $256\times128$ & $206$ \\
            \bottomrule
        \end{tabular}
        \caption{}
        \label{tab:ablat_rec_f}
    \end{subtable}
\end{table}

\begin{table}[t]
    \centering
    \caption{Ablation studies of the key design options in our implicit diffusion model, tested with NeRV-Diffusion-S and the best NeRV autoencoder configuration on UCF.}
    \label{tab:ablat_gen}
    \begin{subtable}[t]{0.45\linewidth}
        \centering
        \begin{tabular}{lc}
            \toprule
            Model & gFVD$\downarrow$ \\
            \midrule
                G.pt \citep{peebles2022learning} & $550$ \\
            DiT \citep{peebles2023scalable} & $\mathbf{295}$ \\
            Latte \citep{ma2024latte} & $342$ \\
            \bottomrule
        \end{tabular}
        \caption{}
        \label{tab:ablat_gen_a}
    \end{subtable}
    \hfill
    \begin{subtable}[t]{0.45\linewidth}
        \centering
        \begin{tabular}{lc}
            \toprule
            Configurations & gFVD$\downarrow$ \\
            \midrule
            Vanilla DiT & $295$ \\
            w/ Min-SNR-$\gamma$ & $238$ \\
            w/ Scheduled Sampling & $261$ \\
            w/ Both & $\mathbf{184}$ \\
            \bottomrule
        \end{tabular}
        \caption{}
        \label{tab:ablat_gen_b}
    \end{subtable}
\end{table}

\subsection{Properties of Generative NeRV}

\subsubsection{Long Video Generation via Time Interpolation and Extrapolation}
\label{sec:t_polate}

Benefited from the continuous frame index positional embedding, our generative NeRV features flexible time interpolation and extrapolation capability.
In Figure \ref{fig:t_polate}, we interpolate the input time embeddings by a factor of $8\times$ to sample $128$-frame videos with smooth and distinct motions.
This property indicates that our generative NeRV efficiently encodes high-density information and understand the residual intrinsic of frame sequences.
It enables compact representation of long videos and efficient training with fewer frames and large frame intervals.

\subsubsection{Generative NeRV Weight Interpolation}

Our generative NeRV also features smooth interpolation between two distinct videos by interpolating their instance parameters.
Given two generative NeRVs' parameters $\theta_{1}$ and $\theta_{2}$, we perform linear interpolation $\lambda\theta_{1}+(1-\lambda)\theta_{2}$ between them.
The visual outcomes are exhibited in Figure \ref{fig:supp_winterp}.
Our model produces progressively interpretable results compared to DIGAN.
We attribute this parametric continuity to not only the Gaussian distribution constraint of our weight latent, but also our simple yet effective linear bottleneck mapping and channel-wise parameterization.

%% file: secs/5_conclusion.tex
\section{Conclusion}

We propose NeRV-Diffusion, a two-staged video synthesis model via NeRV weight generation.
Our NeRV autoencoder projects videos into a Gaussian weight latent space for tokenization, where our implicit diffusion model denoises to generate neural weights that render into videos.
NeRV-Diffusion outperforms both INR-based and most recent non-implicit video generative models on multiple real-world video benchmarks, demonstrating promising scaling law.
It also features smooth temporal and parametric interpolation properties.
The outstanding performance of NeRV-Diffusion highlights the potential of a new holistic video synthesis paradigm with efficient representations.

\section{Acknowledgments}

This work was partially supported by NSF CAREER Award (\#2238769) to AS. The authors acknowledge UMD’s supercomputing resources made available for conducting this research. The U.S. Government is authorized to reproduce and distribute reprints for Governmental purposes notwithstanding any copyright annotation thereon. The views and conclusions contained herein are those of the authors and should not be interpreted as necessarily representing the official policies or endorsements, either expressed or implied, of NSF or the U.S. Government.

%% file: secs/X_suppl.tex
\clearpage
\setcounter{page}{1}

\renewcommand{\thefigure}{A\arabic{figure}}
\renewcommand{\thetable}{A\arabic{table}}
\setcounter{figure}{0}
\setcounter{table}{0}

\section{Implementation Details}
\label{sec:suppl_implem}

\subsection{Additional Model and Training Details}

We use a medium configuration of ViT with $18$ blocks, $14$ heads and $896$ hidden dimensions for our NeRV encoder.
We set the scale of KL divergence loss to $1\times10^{-5}$.
It patchifies the RGB videos into $8\times8\times1$ patches along height, width and time dimensions.
We use sinusoidal positional embedding for our generative NeRV's input time index, instead of the exponential embedding in vanilla NeRV.
For residual connections we use bilinear to upsample the earlier feature maps before merging them into the main branch.

Our discriminator is adapted from a 3D StyleGAN with $5$ blocks, $64$ unit hidden dimensions and a channel multiplier of $2$ for each block.
Its learning rate is set to one fifth of the NeRV autoencoder's, and it is updated every five iterations to stabilize the training.
The scale of the GAN loss added to our NeRV autoencoder is $1$.

Our implicit diffusion transformer adopts DiT-L configuration with $24$ layers, $16$ heads and $1024$ hidden dimensions.
Its patch size follows the token shape as output by our NeRV encoder.
Our implicit DiT is optimized for predicting the noise $\epsilon$ at each timestep, and thus also adjust the Min-SNR-$\gamma$ loss weighting accordingly.
We employ CFG for class-conditioned sampling and the optimal guidance scale is $2$.

Both our NeRV autoencoder and implicit DiT train with L2 reconstruction loss.
We use AdamW\citep{loshchilov2017decoupled} optimizer with a linear warmup learning rate schedule and cosine decay.
Both learning rates are set to $1\times10^{-4}$.
We train our NeRV autoencoder for $2$M iterations and train our implicit DiT for for $1$M iterations.

\subsection{Generative NeRV Architecture}

We list the layer-wise architecture of our NeRV model in Table~\ref{tab:nerv_arch}.
``Feature Map'' refers to the output feature map of  each layer.
``Modulation Weight'' refers to the instance-specific weight latent to be assigned to each NeRV layers.
$T$ is the number of frames.

We set the dimensions of the time, height and width positional embeddings all to $16$.
We start from sampling a spatiotemporal positional embedding of shape $[8, 8, T, 48]$.
It is transposed to queries along the time axis $[T, 48, 8, 8]$, and then spatial convolutions are applied on it.

We use kernel size $k=4$ for all upsampling transposed convolutions and $k=3$ for all other convolutions that don't change the feature map shape.
We set the base hidden dimensions $D=128,256,512$ for NeRV-Diffusion-S, -B and -L configurations, respectively.
$\mathrm{gelu}$ is used for activations in all blocks while $\mathrm{tanh}$ is used after the tailing toRGB layer.

\begin{table}[ht]
    \centering
    \caption{Detailed architecture of our generative NeRV model and modulated weight shape.
    Batch size is omitted.
    }
    \label{tab:nerv_arch}
    \resizebox{0.7\linewidth}{!}{
    \begin{tabular}{c|c|c}
    \toprule
    Layer & \makecell{Feature \\ Map Shape} & \makecell{Modulation \\ Weight Shape} \\
    \midrule
    Input Init & $[8, 8, T, 48]$ & - \\
    Reshape & $[T, 48, 8, 8]$ & - \\
    \midrule
    Conv & $[T, D, 8, 8]$ & $[48, D, 3, 3]$ or $[24, D/2, 3, 3]$ \\
    Transposed Conv & $[T, D, 16, 16]$ & $[64, 64, 4, 4]$ \\
    Conv & $[T, D, 16, 16]$ & $[64, 64, 3, 3]$ \\
    \midrule
    Conv & $[T, D, 16, 16]$ & $[64, 64, 3, 3]$ \\
    Transposed Conv & $[T, D, 32, 32]$ & $[64, 64, 4, 4]$ \\
    Conv & $[T, D, 32, 32]$ & $[64, 64, 3, 3]$ \\
    \midrule
    Conv & $[T, D, 32, 32]$ & $[64, 64, 3, 3]$ \\
    Transposed Conv & $[T, D, 64, 64]$ & $[64, 64, 4, 4]$ \\
    Conv & $[T, D, 64, 64]$ & $[64, 64, 3, 3]$ \\
    \midrule
    Conv & $[T, 2D, 64, 64]$ & $[128, 64, 3, 3]$ \\
    Transposed Conv & $[T, 2D, 128, 128]$ & $[64, 64, 4, 4]$ \\
    Conv & $[T, 2D, 128, 128]$ & $[64, 64, 3, 3]$ \\
    \midrule
    toRGB & $[T, 3, 128, 128]$ & $[D, 3, 3, 3]$ \\
    Reshape to Output & $[T, 128, 128, 3]$ & - \\
    \bottomrule
    \end{tabular}
    }
\end{table}

\section{Time Interpolation}

\begin{figure}[t]
    \centering
    \includegraphics[width=\linewidth]{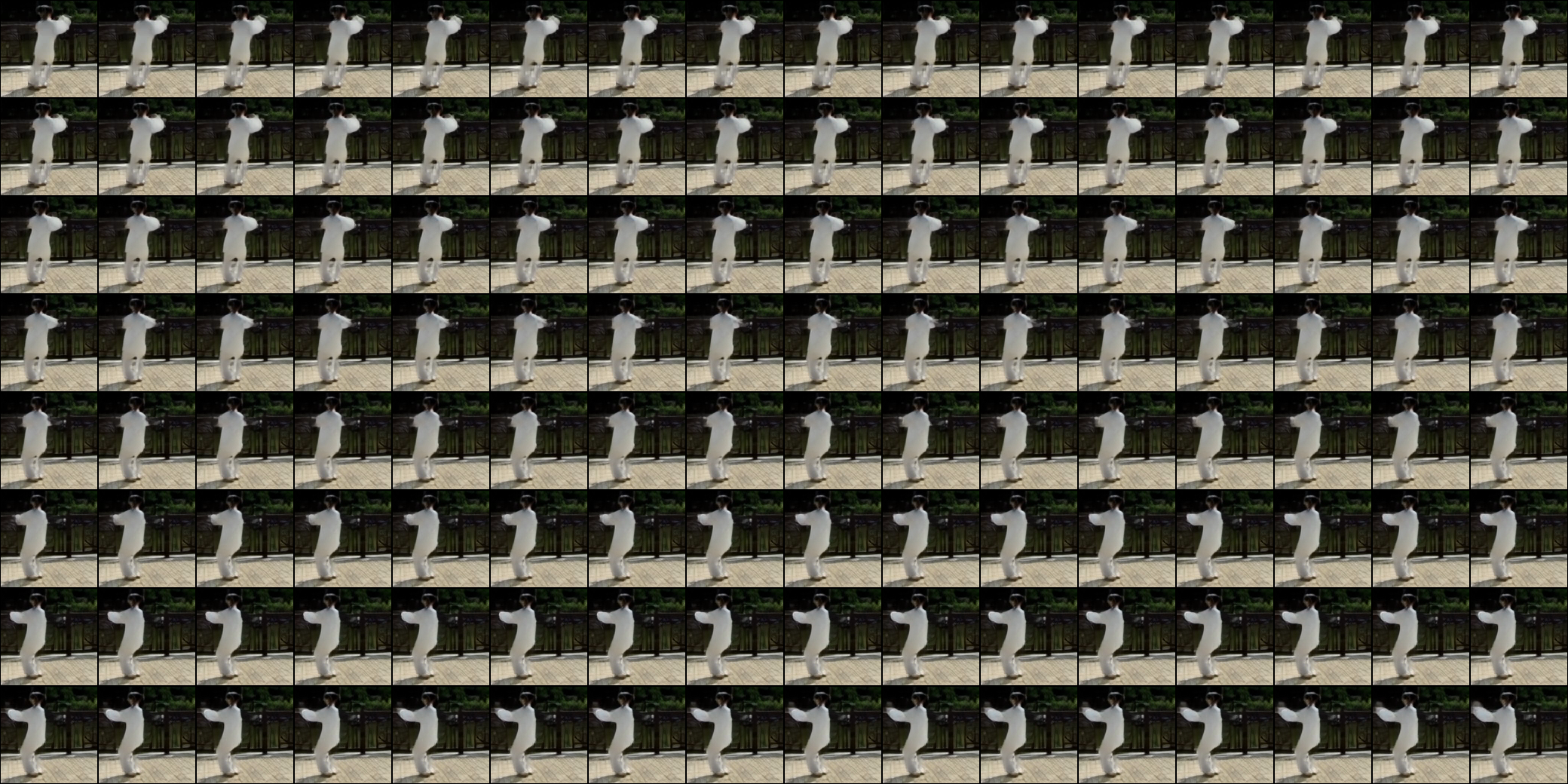}
    \par\vspace{3mm}
    \includegraphics[width=\linewidth]{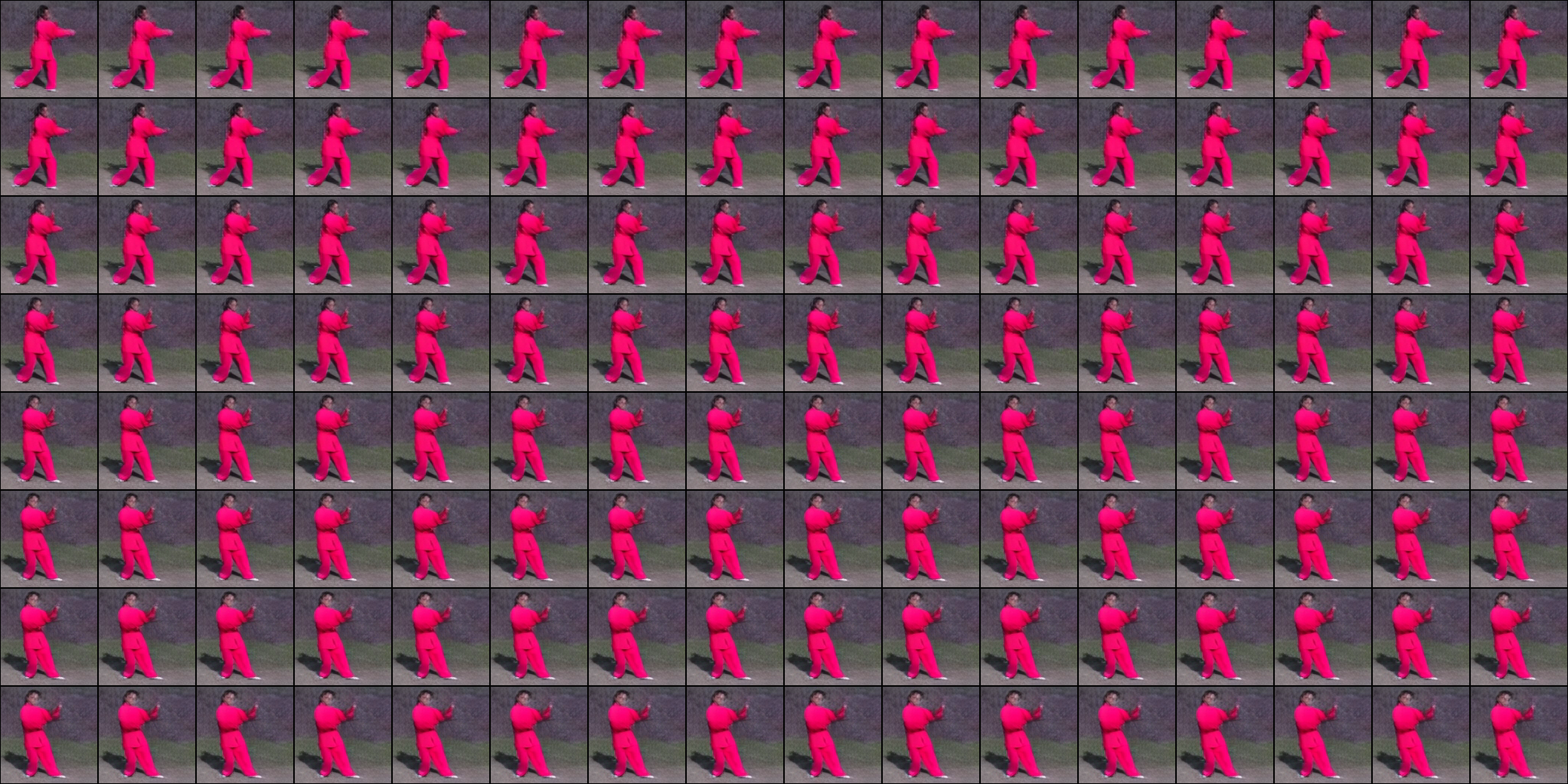}
    \caption{128-frame video interpolation results.
    NeRV-Diffusion can be easily extended to efficient long video generation via smooth time interpolation after trained with large frame intervals.
    }
    \label{fig:t_polate}
\end{figure}

As discussed in \S\ref{sec:t_polate}, we train NeRV-Diffusion on TaiChi-HD \citep{siarohin2019first} dataset with an interval of $4$ frames, and interpolate the input time embeddings to sample 128-frame videos.
The results are presented in Figure \ref{fig:t_polate}.

\section{INR Weight Interpolation}
\label{sec:suppl_winterp}

\begin{figure}[t]
    \centering
    \includegraphics[width=0.8\linewidth]{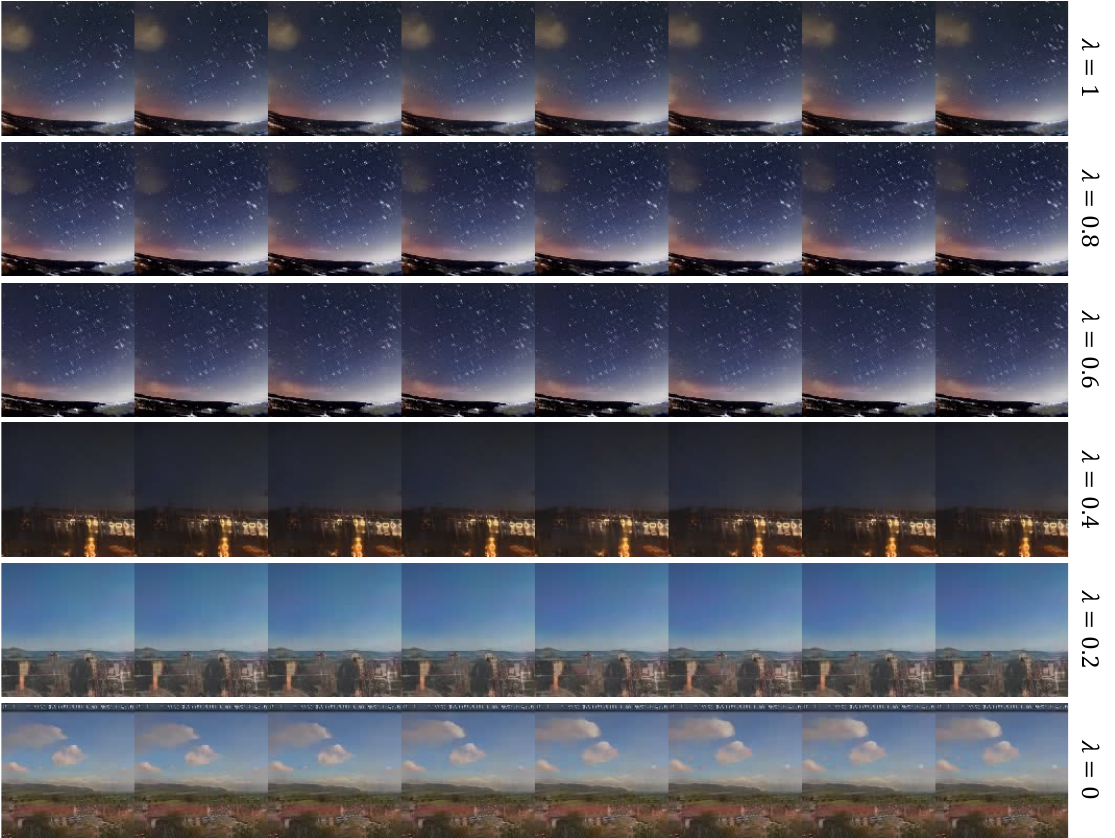}
    \par\vspace{3mm}
    \includegraphics[width=0.8\linewidth]{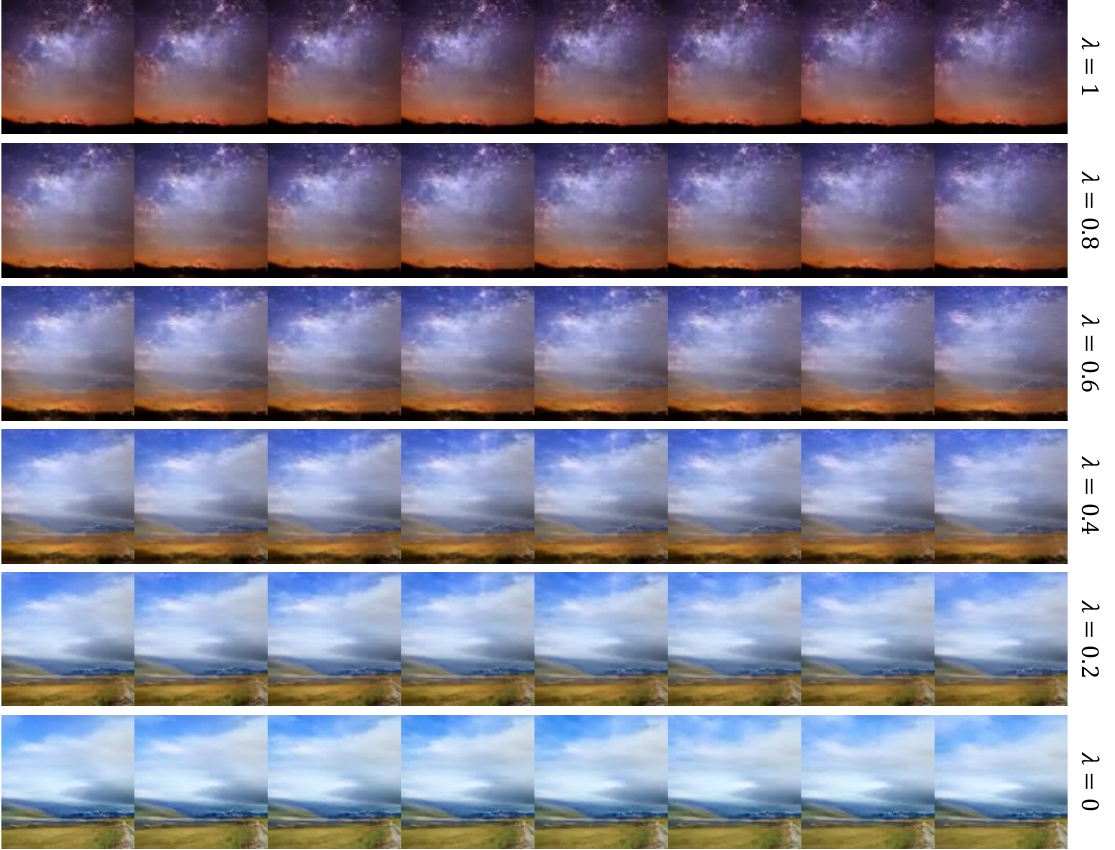}
    \caption{Interpolation of the whole parameters of the video INRs in DIGAN~\citep{yu2022generating} (top) and our generative NeRVs (bottom).
    }
    \label{fig:supp_winterp}
\end{figure}

We further illustrate our generative NeRV's superiority in INR weight interpolation.
DIGAN \citep{yu2022generating} proposes to interpolate between the latent noise vectors.
When being interpolated between the whole weights, their video INR presents non-continuous transitions as shown in Figure \ref{fig:supp_winterp} (top).
This is because
1) their latent vectors are decoded from Gaussian noise with a complex non-linear mapping network;
2) their INR weights are modulated with low-rank cross product, termed as Factorized Matrix Multiplication (FMM) in \cite{skorokhodov2021adversarial}), of the latent vectors, which break the arithmetic property.
In contrast, our weight latent is directly used for modulation with a single linear affine layer from the KL bottleneck, and is directly assigned as NeRV parameters with minimal transforms.
Our generative NeRV presents smooth interpolation effect as shown in Figure \ref{fig:supp_winterp} (bottom).
This property also opens up the potential of general direct manipulations on the tokenized NeRVs in a compositional manner with our NeRV autoencoder.

\section{Additional Qualitative Results}
\label{sec:supp_exp}

We provide more generation samples of NeRV-Diffusion on UCF dataset in Figure \ref{fig:supp_ucf_c}.
Video files in MP4 format are attached in the supplementary materials.

\begin{figure}[t]
    \centering
    \includegraphics[width=\linewidth]{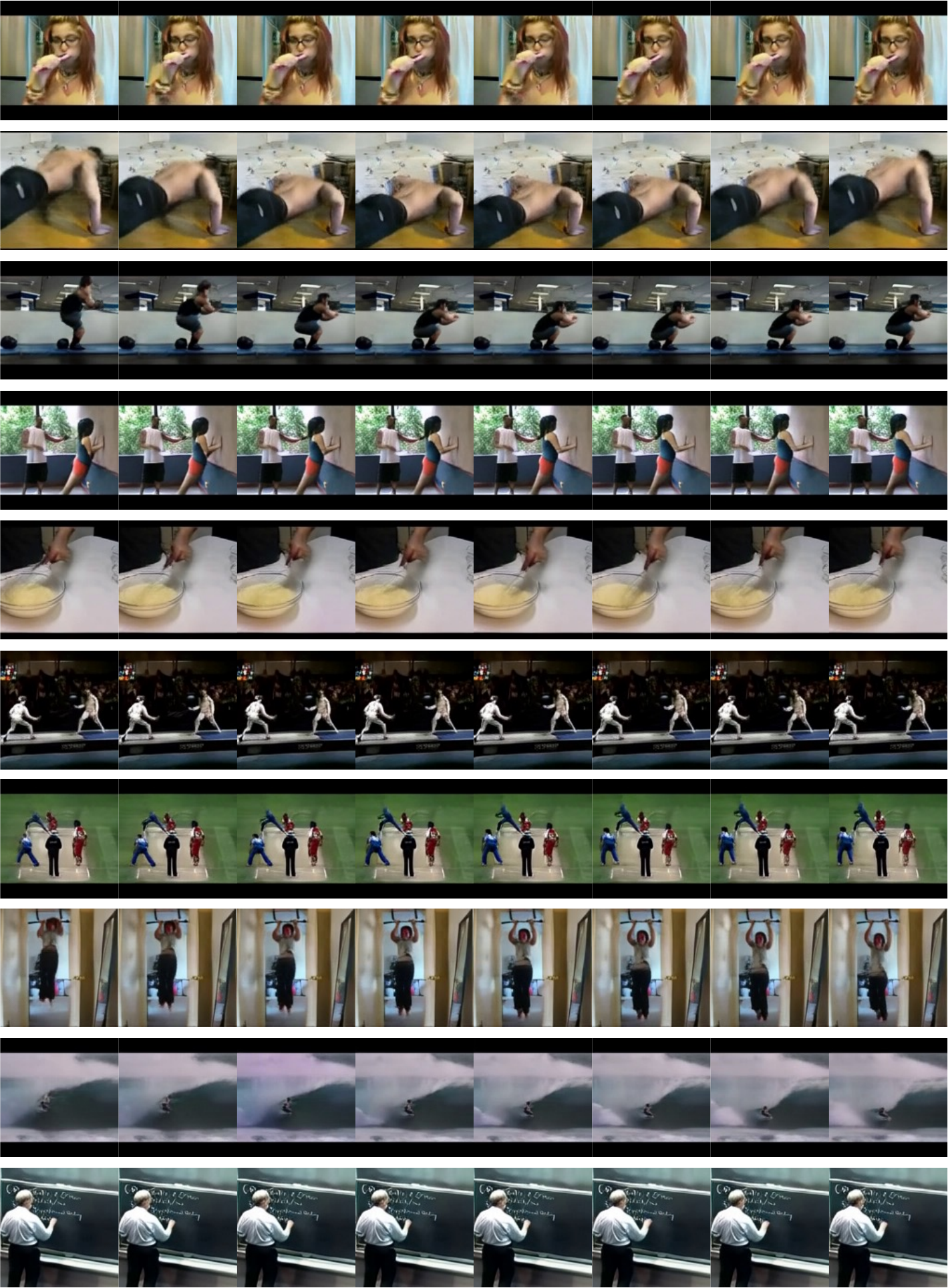}
    \caption{Additional class-conditioned generation samples on UCF.
    }
    \label{fig:supp_ucf_c}
\end{figure}